\documentclass{article}
\usepackage{spconf,amsmath,graphicx}%
\usepackage{booktabs}
\usepackage{graphicx}

\usepackage{enumitem}
\setlist{nosep, leftmargin=14pt}

\usepackage{mwe} 

\title{Domain-Expert-Guided Hybrid Mixture-of-Experts for Medical AI: Integrating Data-Driven Learning with Clinical Priors}
\name{
Jinchen Gu$^{1}$\,,
Nan Zhao$^{1}$\footnotemark[1]\sthanks{These authors contributed equally as second authors.},
Lei Qiu$^{1}$\footnotemark[1],
Lu Zhang$^{1}$\sthanks{Corresponding author: lz50@iu.edu.}
}

\address{
$^{1}$ Department of Computer Science, Indiana University Indianapolis, Indianapolis, IN, USA
}
\begin{document}
%
\maketitle
\begin{abstract}
Mixture-of-Experts (MoE) models increase representational capacity with modest computational cost, but their effectiveness in specialized domains such as medicine is limited by small datasets. In contrast, clinical practice offers rich expert knowledge, such as physician gaze patterns and diagnostic heuristics, that models cannot reliably learn from limited data. Combining data-driven experts, which capture novel patterns, with domain-expert-guided experts, which encode accumulated clinical insights, provides complementary strengths for robust and clinically meaningful learning. To this end, we propose Domain-Knowledge-Guided Hybrid MoE (DKGH-MoE), a plug-and-play and interpretable module that unifies data-driven learning with domain expertise. DKGH-MoE integrates a data-driven MoE to extract novel features from raw imaging data, and a domain-expert-guided MoE incorporates clinical priors, specifically clinician eye-gaze cues, to emphasize regions of high diagnostic relevance. By integrating domain expert insights with data-driven features, DKGH-MoE improves both performance and interpretability. (Code: https://github.com/BrainX-Lab/Domain-Expert-Guided-Hybrid-MoE)
\end{abstract}

\section{Introduction}
\label{sec:intro}
\begin{figure*}[!t]
  \centering
  \includegraphics[width=.8\textwidth]{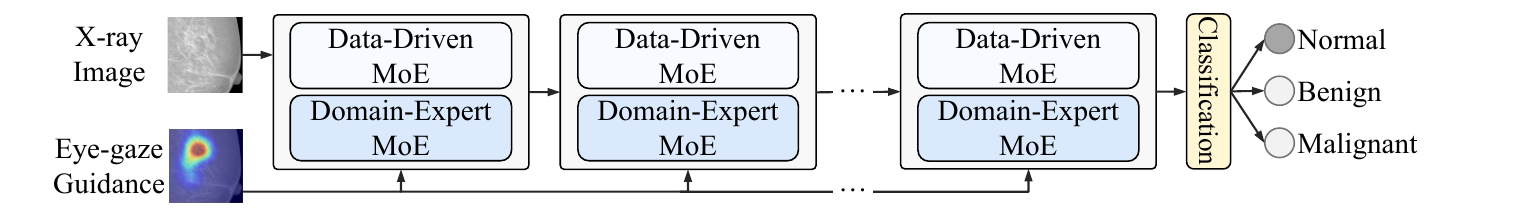}
\vspace{-0.3cm}
\caption{Overall architecture of the network with the plug-and-play DKGH-MoE module incorporated.}
\label{fig:overview}
\vspace{-.22cm}
\end{figure*}

Mixture-of-Experts (MoE) has emerged as a powerful paradigm in modern machine learning, enabling models to scale parameter capacity through conditional computation by activating a subset of experts for each input. This mechanism allows MoE models to achieve substantially enhanced representational capacity with relatively modest computational overhead, and has driven significant advances in large-scale foundation models \cite{cai2025survey}. Prior work has shown that MoE models benefit greatly from large-scale datasets and high expert diversity, which together promote meaningful expert specialization and strong generalization across tasks \cite{du2022glam}.

However, large-scale datasets are rarely satisfied in specialized domains such as healthcare. Clinical datasets are typically small, costly to acquire, and highly heterogeneous across institutions and patient populations \cite{guan2021domain}. Under these constraints, purely data-driven MoE models struggle to develop robust and semantically meaningful expert specialization, resulting in only marginal gains in existing medical applications \cite{jiang2024m4oe,xiang2025mixture} compared to their success in general-domain settings \cite{du2022glam}. This gap underscores a fundamental mismatch between the data-intensive nature of MoE architectures and the realities of clinical AI. While prior medical imaging studies have incorporated expert modules or attention mechanisms guided by anatomical priors or task-specific heuristics, they do not explicitly integrate clinician knowledge into MoE routing or expert specialization \cite{Eye_Gaze_ma2024eye,chen2024low}.

In contrast to data scarcity, clinical domains possess a wealth of domain knowledge accumulated over decades, such as clinician gaze patterns, diagnostic heuristics, lesion localization strategies, and well-established clinical priors \cite{gao2021medical}. Such expert knowledge encodes high-level reasoning cues that are difficult for models to infer directly from limited data. Effectively integrating this knowledge into model learning can: (1) reduce the data requirements for expert specialization; (2) guide models toward clinically meaningful features that may be subtle or rare, and (3) promote a learning paradigm aligned with human cognition, where new knowledge builds upon prior expert insights. In this sense, combining clinical priors with data-driven learning enables models to ``stand on the shoulders of giants`` rather than rediscovering foundational patterns from scratch.

To address these challenges, we propose a Domain-Knowledge-Guided Hybrid Mixture-of-Experts (DKGH-MoE), a plug-and-play and interpretable MoE module that integrates data-driven experts with domain-specific experts, as illustrated in Fig.~\ref{fig:overview}. The hybrid architecture combines a data-driven MoE that discovers novel patterns directly from imaging data with a domain-expert-guided MoE that incorporates clinical priors, specifically clinician eye-gaze cues in this work, to emphasize regions of high diagnostic relevance. These expert signals inform the routing mechanism by grouping image regions with similar clinical significance to the same expert, facilitating meaningful expert specialization even under limited data. By unifying accumulated expert knowledge with learned representations, DKGH-MoE improves both performance and interpretability of MoE-based medical imaging models.

\section{Methods}
\label{sec:methods}

The proposed DKGH-MoE is a plug-and-play module that can be seamlessly integrated into existing neural network backbones by replacing standard layers with DKGH-MoE blocks. This section details the module components, including the Data-Driven MoE, the Domain-Expert MoE, the MoE fusion mechanism, and the associated training objectives.

\subsection{Module Structure}
As shown in Fig.~\ref{fig:Hybrid_MoE}, the module consists of three main parts: (1) a Data-Driven MoE (DD-MoE) that learns only from image features; (2) a Domain-Expert MoE (DE-MoE) that leverages clinical expert eye-gaze features to guide expert selection; and (3) a MoE Fusion module that combines outputs from both branches.

\subsubsection{Data-Driven MoE}
\begin{figure}[h]
  \centering
  \includegraphics[width=6.5cm]{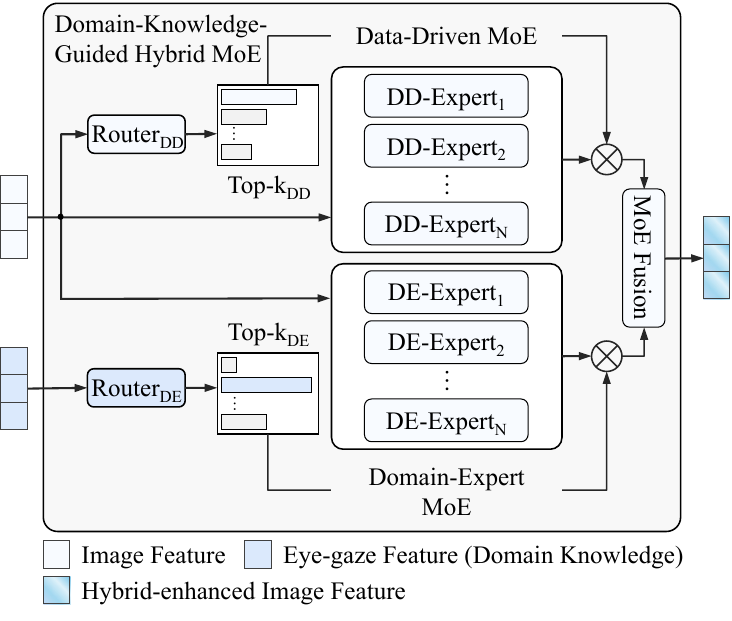}
\vspace{-0.2cm}
\caption{Architecture of DKGH-MoE module.}
\label{fig:Hybrid_MoE}
\vspace{-.3cm}
\end{figure}

The Data-Driven MoE (DD-MoE) follows a standard MoE design and consists of a router network and $N$ independent DD-expert modules. As illustrated in Fig.~\ref{fig:Hybrid_MoE}, the input image feature
$x$ is first transformed into a compact representation $x_f$ through global average pooling, which is used for routing. The router network, implemented as a lightweight MLP, takes $x_f$ as input and produces expert scores, $r^{(\mathrm{DD})}$. These scores determine which experts should be activated, and only the $\mathrm{Top\text{-}K}$ experts with highest scores are selected for computation, enabling sparse activation and efficient parameter utilization. Each DD-expert $E_i^{(\mathrm{DD})}$ shares the same architecture as the backbone block it replaces, but maintains its own parameters to allow expert specialization. The output of DD-MoE is obtained by a weighted sum over the selected experts:
\begin{equation}
\setlength{\abovedisplayskip}{4pt}
\setlength{\belowdisplayskip}{4pt}
\textstyle
  h^{(\mathrm{DD})} = \sum_{i\in \mathrm{Top\text{-}K}(r^{(\mathrm{DD})})} \mathrm{softmax}\!\big(r_i^{(\mathrm{DD})}(x_f)\big) \cdot E_i^{(\mathrm{DD})}(x),
\end{equation}
where the softmax-normalized router scores serve as combination weights, ensuring that the routing decision reflects both the input features and the relative importance assigned to each expert. This mechanism enables the DD-MoE to learn diverse data-driven representations and encourages expert specialization, while maintaining computational efficiency through sparse activation. 

\subsubsection{Domain-Expert MoE}
The DE-MoE is designed to incorporate clinical domain knowledge directly into the expert routing process. Its overall structure mirrors that of the DD-MoE, but the DE-MoE router uses domain-expert features to guide expert selection. Formally, the DE-MoE output is computed as:
\begin{equation}
\setlength{\abovedisplayskip}{4pt}
\setlength{\belowdisplayskip}{4pt}
\textstyle
  h^{(\mathrm{DE})} = \sum_{ i\in \mathrm{Top\text{-}K}(r^{(\mathrm{DE})})} \mathrm{softmax}\!\big(r_i^{(\mathrm{DE})}(x_{\mathrm{exp}})\big) \cdot E_i^{(\mathrm{DE})}(x),
\end{equation}
where $x_{\mathrm{exp}}$ denotes the domain-expert feature extracted by feeding the clinician eye-tracking heatmap into a convolutional network. These gaze-derived features provide a meaningful proxy for clinical attention patterns: although gaze alone does not determine the final diagnosis, it highlights regions that clinicians typically inspect during decision-making. By conditioning the routing on $x_{\mathrm{exp}}$, images with similar expert-attended regions are encouraged to activate the same subset of DE-experts. This promotes expert specialization around clinically informative cues and encourages each expert to focus on patterns aligned with real-world diagnostic strategies. As a result, the DE-MoE strengthens the interpretability of the learned features and enhances the alignment between model behavior and clinical reasoning.

\subsubsection{MoE Fusion Module}
To enhance the robustness and adaptability of the overall framework and to avoid relying solely on a single routing mechanism, particularly given the potential noise in expert-guided signals, we fuse the outputs of the DD-MoE and DE-MoE using a probabilistic gating function. The fused representation is computed as:
\begin{equation}
\setlength{\abovedisplayskip}{4pt}
\setlength{\belowdisplayskip}{4pt}
\textstyle
  \hat{x} = p \cdot h^{(\mathrm{DE})} + (1-p) \cdot h^{(\mathrm{DD})},
\end{equation}
where $\hat{x}$ denotes the final output of the DKGH-MoE module. The gate $p$ adaptively balances the contributions of the expert-guided and data-driven branches:
\begin{equation}
\setlength{\abovedisplayskip}{4pt}
\setlength{\belowdisplayskip}{4pt}
\textstyle
  p = \sigma\big(w_p(x_f, x_{\mathrm{exp}})\big),
\end{equation}
where $w_p$ is a linear layer and $\sigma$ is the sigmoid function to keep $p$ in $[0,1]$. Conditioning the gate on both the image-derived feature $x_f$ and the expert-derived feature $x_{\mathrm{exp}}$ allows the module to dynamically adjust its reliance on domain expertise versus data-driven learning, improving stability under noisy expert signals and enhancing overall model robustness.

\subsection{Loss Function}
The overall training objective consists of a classification loss and a routing load-balancing loss. The classification loss is cross-entropy between predictions and ground-truth labels:
\begin{equation}
\setlength{\abovedisplayskip}{4pt}
\setlength{\belowdisplayskip}{4pt}
\textstyle
  \mathcal{L}_{\mathrm{cls}}=\mathrm{CE}(y_{\mathrm{pred}},y_{\mathrm{true}}),
\end{equation}
To encourage all experts to participate in routing and prevent the model from overusing only a small subset of experts, we follow \cite{lepikhin2020gshard} and adopt a load-balancing term:
\begin{equation}
\setlength{\abovedisplayskip}{4pt}
\setlength{\belowdisplayskip}{4pt}
\textstyle
  \mathcal{L}_{\mathrm{lb}}=\sum_{i=1}^{N} f_i\, p_i,
\end{equation}
where ${N}$ is the total number of experts, $f_i$ is the usage frequency of the $i$-th expert, and $p_i$ is the average routing probability assigned to that expert. This loss promotes more uniform expert utilization during training.
The final objective is:
\begin{equation}
\setlength{\abovedisplayskip}{4pt}
\setlength{\belowdisplayskip}{4pt}
\textstyle
\mathcal{L}=\mathcal{L}_{\mathrm{cls}}+\lambda\,\mathcal{L}_{\mathrm{lb}}.
\end{equation}
where we set $\lambda=0.01$ to balance classification performance and routing regularization.

\section{Experiments}
\label{sec:experiments}
\setlength{\tabcolsep}{4.4pt}  
\begin{table*}[!t]
\centering
\small
\caption{Quantitative results on ResNet-18 and ResNet-50 backbones (\%).}
\label{tab:quan_res}
\begin{tabular}{l c c c c c c c c}
\toprule
 & \multicolumn{4}{c}{ResNet-18} & \multicolumn{4}{c}{ResNet-50} \\
\cmidrule(lr){2-5} \cmidrule(lr){6-9}
\textbf{Methods} & \multicolumn{2}{c}{Dense} & \multicolumn{2}{c}{Sparse} & \multicolumn{2}{c}{Dense} & \multicolumn{2}{c}{Sparse} \\
\cmidrule(lr){2-3} \cmidrule(lr){4-5} \cmidrule(lr){6-7} \cmidrule(lr){8-9}
 & ACC & AUC & ACC & AUC & ACC & AUC & ACC & AUC \\
\hline
baseline & 62.92$\pm$10.18 & 70.45$\pm$12.15 & --- & --- & 56.00$\pm$12.28 & 67.42$\pm$4.27 & --- & --- \\
DD-MoE & 70.45$\pm$12.15 & 72.46$\pm$14.04 & 62.62$\pm$12.27 & 68.10$\pm$12.13 & 54.85$\pm$10.41 & 61.46$\pm$4.54 & 57.04$\pm$6.19 & 59.02$\pm$9.52 \\
DE-MoE & 65.02$\pm$9.57 & 72.03$\pm$10.78 & 72.77$\pm$4.10 & 68.88$\pm$9.47 & 58.20$\pm$9.31 & 58.46$\pm$4.92 & 66.38$\pm$5.73 & 62.50$\pm$8.63 \\
DKGH-MoE & \textbf{77.79$\pm$10.14} & \textbf{82.91$\pm$7.50} & \textbf{79.12$\pm$7.59} & \textbf{80.84$\pm$10.56} & \textbf{77.23$\pm$8.54} & \textbf{84.01$\pm$7.76} & \textbf{75.46$\pm$8.85} & \textbf{81.07$\pm$7.77} \\
\bottomrule
\end{tabular}
\end{table*}

\subsection{Dataset and Implementation Details}

\noindent\textbf{Implementation Details:} We integrate DKGH-MoE into ResNet-18 and ResNet-50 backbones by replacing residual blocks with DKGH-MoE blocks.  Each Data-Driven MoE (DD-MoE) and Domain-Expert MoE (DE-MoE) uses $N=4$ experts, where each expert mirrors the structure of the replaced block (basic blocks for ResNet-18 and bottleneck blocks for ResNet-50) \cite{he2016deep}. We evaluate both dense routing ($k=N$) and sparse routing ($k=1$). Models are trained using Adam with a learning rate of $5\times 10^{-4}$ and a StepLR scheduler. Batch sizes are 64 for ResNet-18 and 16 for ResNet-50. Experiments are conducted on a single NVIDIA H100 GPU.

\noindent\textbf{Dataset:} INBreast consists of 410 mammography images from 115 patients \cite{moreira2012inbreast}, annotated with BI-RADS assessments and labeled as normal, benign, or malignant. Eye-gaze data are collected and processed following \cite{ma2023eye}, to generate gaze heatmaps aligned with each image. Images are cropped to $1024\times 1024$: around mass masks for mass cases and randomly for non-mass cases. We adopt five-fold subject-wise cross-validation (80\% training / 20\% testing). During training, for data augmentation, brightness/contrast are adjusted within $[0.8,1.2]$, gaussian noise ($\sigma=0.05$) is added, and uniform class sampling is used to mitigate class imbalance.

\subsection{Quantitative Analysis}
We evaluate the effectiveness of the proposed module by comparing four variants: a plain ResNet baseline, DD-MoE, DE-MoE, and DKGH-MoE. Both dense MoE (Top-k = $N$) and sparse MoE (Top-k = 1) are evaluated, with quantitative results summarized in Table~\ref{tab:quan_res}. 

\textbf{ResNet-18.} Under the dense setting, DD-MoE improves accuracy (ACC) from 62.92\% → 70.45\% and area under the curve (AUC) from 70.45\% → 72.46\%, demonstrating the benefit of data-driven expert specialization. DE-MoE also outperforms the baseline but shows less stability, likely due to noise in gaze-derived signals. DKGH-MoE achieves the best performance - 77.79\% ACC and 82.91\% AUC, confirming the complementary strengths of data-driven and expert-guided routing. Under sparse routing, DKGH-MoE again performs best (79.12\% ACC, 80.84\% AUC), matching dense MoE performance with far fewer activated experts, highlighting both efficiency and preserved learning capacity.

\textbf{ResNet-50.} For the deeper backbone, adding a generic MoE module (DD-MoE or DE-MoE alone) does not consistently improve performance and can even degrade it, as ResNet-50 already has strong representational capacity and additional MoE complexity increases overfitting risk. In contrast, DKGH-MoE consistently yields substantial gains, achieving 77.23\% ACC / 84.01\% AUC under dense setting and 75.46\% ACC / 81.07\% AUC under the sparse setting. 

Overall, DKGH-MoE delivers the strongest performance across all backbones and routing settings, demonstrating the robustness and complementary benefits of integrating data-driven and domain-expert-guided routing.

\subsection{Visualization Analysis}
\begin{figure}[t]
  \centering
  \begin{minipage}[b]{.92\linewidth}
    \centering
    \includegraphics[width=.9\linewidth, trim=0cm 5.5cm 0cm 5.5cm,
  clip]{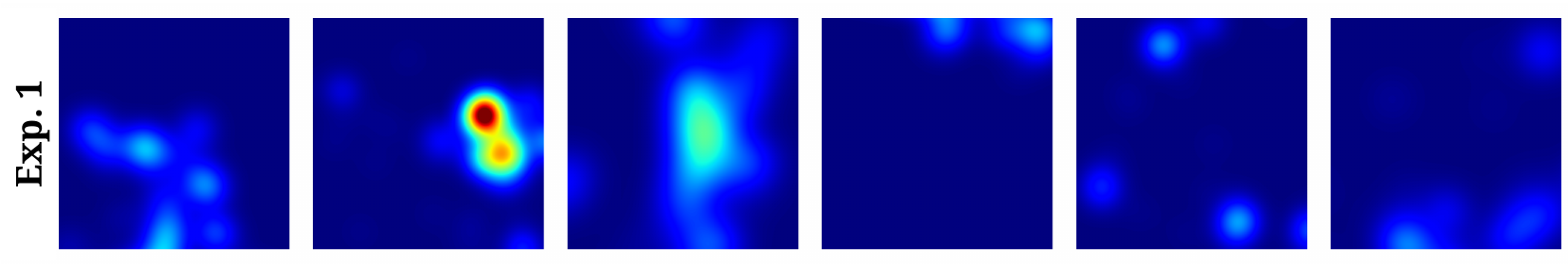}
  \end{minipage}
  
  \vspace{-0.3em}
  \begin{minipage}[b]{.92\linewidth}
    \centering
   \includegraphics[width=.9\linewidth, trim=0cm 5.5cm 0cm 5.5cm,
  clip]{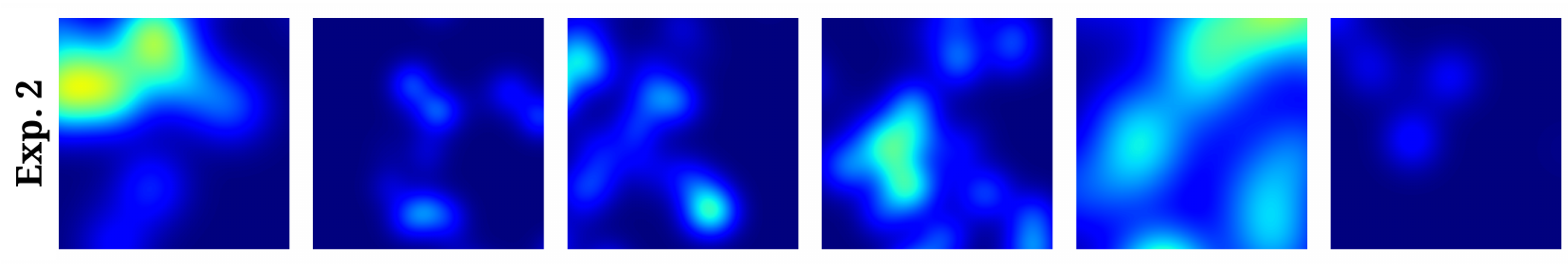}
  \end{minipage}
  
  \vspace{-0.3em}
  \begin{minipage}[b]{.92\linewidth}
    \centering
    \includegraphics[width=.9\linewidth, trim=0cm 5.5cm 0cm 5.5cm,
  clip]{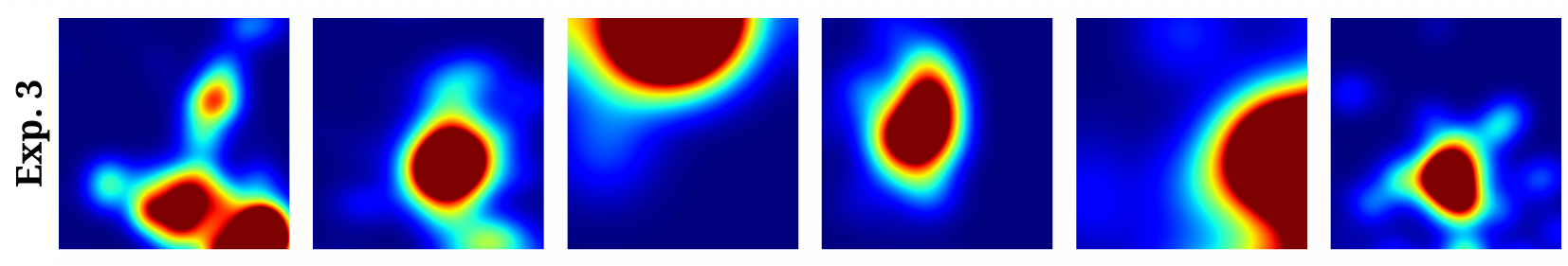}

  \end{minipage}
  
  \vspace{-0.3em}
  \begin{minipage}[b]{.92\linewidth}
    \centering
    \includegraphics[width=.9\linewidth, trim=0cm 5.5cm 0cm 5.5cm,
  clip]{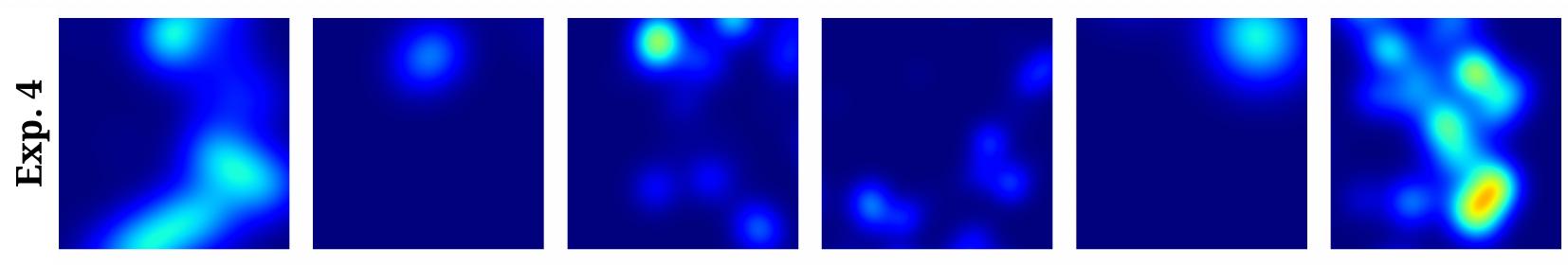}
  \end{minipage}
  
  \vspace{-0.2cm}
  \caption{Top-1 expert assignments. Each row shows six randomly sampled gaze heatmaps routed to the same expert, illustrating expert specialization driven by clinician attention.}
  \label{fig:gaze_groups}
  \vspace{-0.3cm}
\end{figure}

Under the sparse setting ($k=1$), we analyze how DE-MoE assigns image patches to experts and whether these assignments align with gaze-derived attention patterns. Fig.~\ref{fig:gaze_groups} shows representative patches routed to Experts 1-4 in the second DE-MoE module. Clear expert specialization emerges: Expert 1 primarily processes patches with weak or diffuse gaze signals; Expert 2 handles moderately strong and spatially continuous gaze responses; Expert 3 focuses on patches with large, high-intensity gaze regions, indicating a strong clinician attention; and Expert 4 captures residual patterns with minimal gaze relevance. These results demonstrate that the DE-MoE router effectively leverages clinician gaze cues to group patches by clinical attention, enhancing interpretability and revealing complementary expert specialization guided by real clinician behavior.

\section{Conclusion}
Our results demonstrate the value of integrating domain expertise into MoE architectures for medical image analysis. By incorporating clinician eye-gaze patterns as expert guidance, the proposed module enhances both representation learning and interpretability. This highlights the value of combining data-driven learning with domain-expert-derived priors. Future work will extend this framework to additional medical tasks and explore other forms of expert knowledge beyond eye-gaze cues.


\section{Compliance with ethical standards}
\label{sec:ethics}
This research study was conducted retrospectively using human subject data made available in open access by INbreast datasets \cite{moreira2012inbreast}. Ethical approval was not required as confirmed by the license attached with the open access data.

\bibliographystyle{IEEEbib}
\bibliography{strings,refs}

\end{document}